\documentclass[10pt,twocolumn,letterpaper]{article}

\usepackage{iccv}              % To produce the CAMERA-READY version
\newcommand{\PAR}[1]{\vskip2pt \noindent{\bf #1~}}
\usepackage[accsupp]{axessibility}  % Improves PDF readability for those with disabilities.
\usepackage{mathtools} % for improved cases in equations
\usepackage{graphicx}
\usepackage{tabularray}
\usepackage{multirow}
\usepackage{microtype}
\usepackage{siunitx}
\usepackage{xparse}
\usepackage{fp}
\usepackage{float}

\DeclareMathOperator{\sign}{sign}

\definecolor{darkgreen}{RGB}{0,100,0}
\newcommand{\gss}[1]{\textcolor{darkgreen}{\scriptsize{#1}}}
\newcommand{\rss}[1]{\textcolor{red}{\scriptsize{#1}}}

\definecolor{iccvblue}{rgb}{0.21,0.49,0.74}
\usepackage[pagebackref,breaklinks,colorlinks,allcolors=iccvblue]{hyperref}

\def\paperID{14193} % *** Enter the Paper ID here
\def\confName{ICCV}
\def\confYear{2025}

\title{RIPE: Reinforcement Learning on Unlabeled Image Pairs for Robust Keypoint Extraction}

\author{Johannes Künzel$^{1,2}$
\quad
Anna Hilsmann$^1$
\quad
Peter Eisert$^{1,2}$
\\
{\normalsize $^1$Fraunhofer Heinrich-Hertz-Institut, HHI, Germany,
$^2$Humboldt University Berlin, Germany,
}
}

\begin{document}
\maketitle
\begin{abstract}
We introduce RIPE, an innovative reinforcement learning-based framework for weakly-supervised training of a keypoint extractor that excels in both detection and description tasks. 
In contrast to conventional training regimes that depend heavily on artificial transformations, pre-generated models, or 3D data, RIPE requires only a binary label indicating whether paired images represent the same scene.
This minimal supervision significantly expands the pool of training data, enabling the creation of a highly generalized and robust keypoint extractor. 

RIPE utilizes the encoder's intermediate layers for the description of the keypoints with a hyper-column approach to integrate information from different scales.  
Additionally, we propose an auxiliary loss to enhance the discriminative capability of the learned descriptors.

Comprehensive evaluations on standard benchmarks demonstrate that RIPE simplifies data preparation while achieving competitive performance compared to state-of-the-art techniques, marking a significant advancement in robust keypoint extraction and description.
To support further research, we have made our code publicly available at \url{https://github.com/fraunhoferhhi/RIPE}.
\end{abstract}    
\section{Introduction}
\label{sec:introduction}

\begin{figure}[t]
    \centering
    \includegraphics[width=0.8\linewidth]{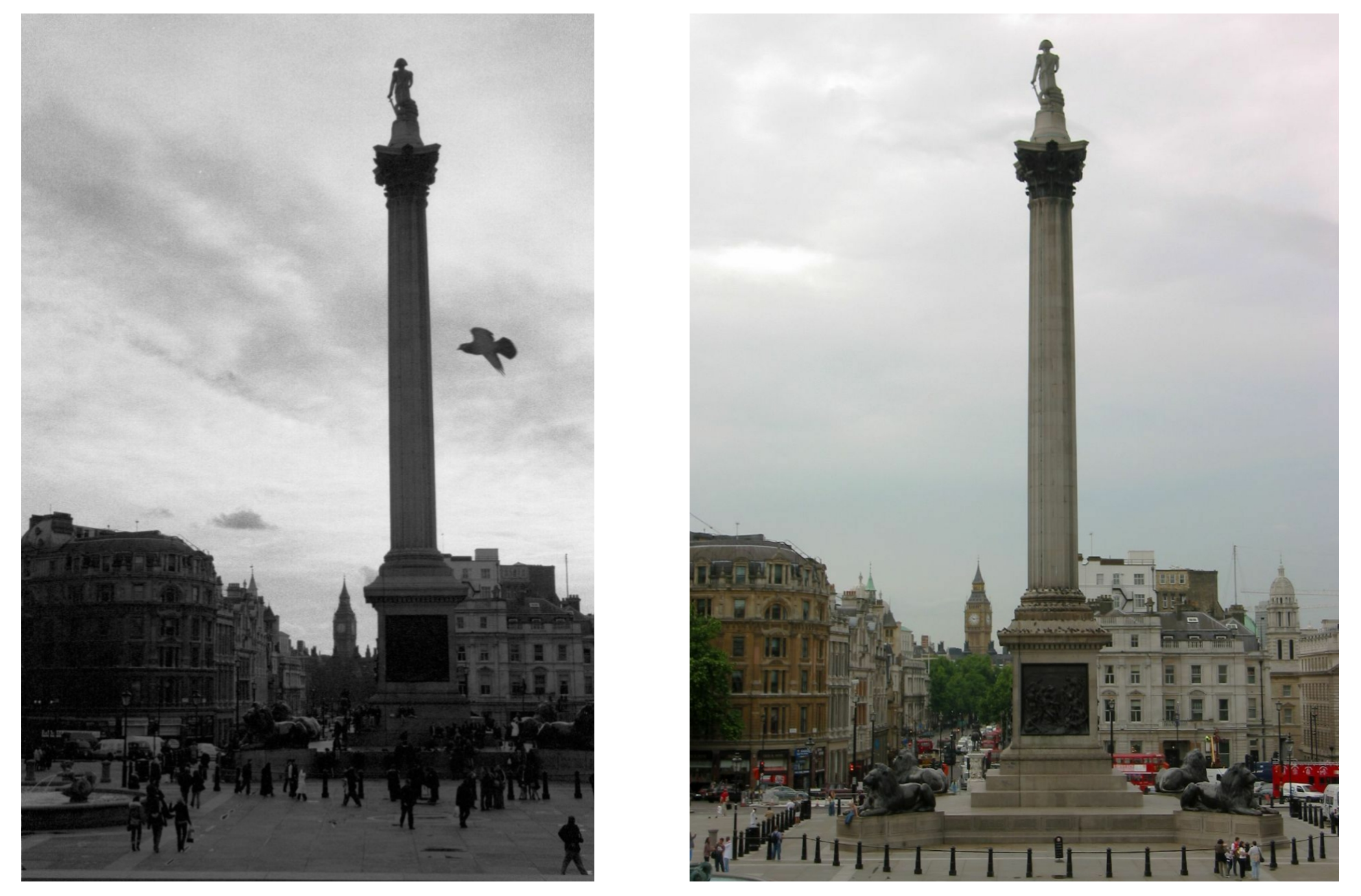}
    \vspace{-0.25cm}
    \caption{Can you tell if these images depict the same scene? While humans naturally ignore noise and lighting variations to solve this effortlessly, teaching a neural network to do the same using only image pairs poses significant challenges. (Images: MegaDepth~\cite{Li2018})}
    \label{fig:teaser}
    \vspace{-0.65cm}
\end{figure}

Given two images, how can we determine whether they depict the same scene and precisely identify matching keypoints?
This task, as shown in \cref{fig:teaser}, is intuitive for humans: we can identify distinctive keypoints in one image and look for their counterparts in the other, ignoring distractions such as moving cars, changing foliage, all while remaining unaffected by noise or lighting variations. 
This natural human capability raises an intriguing question: can neural networks learn robust keypoint extraction exclusively from binary labels indicating whether two images depict the same scene?

Traditional keypoint detection methods such as SIFT~\cite{Lowe.2004}, ORB~\cite{Rublee.2011}, and SURF~\cite{Bay.2006} rely on handcrafted feature detectors and descriptors that struggle significantly with long-term registration tasks, particularly when images are taken hours, days or months apart.
Variations in weather, lighting conditions and appearance (most notably from vegetation) pose significant problems for traditional detectors. 
To overcome these limitations, deep-learning-based methods were introduced to directly learn feature representations from images.
Current state of the art (SOTA) approaches, including DeDoDe~\cite{Edstedt.2024}, DISK~\cite{Tyszkiewicz.2020} and ALIKED~\cite{Zhao.2023dvq} rely on training datasets like MegaDepth~\cite{Li2018}, which provide relative pose and depth information.
MegaDepth, for example, consists of touristic imagery along with 3D models generated with Structure from Motion using COLMAP~\cite{schoenberger2016sfm}, itself reliant on classical SIFT keypoints.
Other methods, like SuperPoint~\cite{DeTone.2018} or SiLK~\cite{Gleize.2023} rely on self-supervision through applying artificial augmentations, but face limitations from the available training data and the domain-gap between real and simulated scenarios, which inadequately replicate long-term changes of the real world.

In this paper we propose RIPE (\textbf{R}einforcement Learning on \textbf{I}mage \textbf{P}airs for Keypoint \textbf{E}xtraction), a novel approach that trains keypoint detectors using only image pairs labeled as either showing the same image scene or not, eliminating the need for depth or pose information.
This significantly expands the pool of usable datasets to include diverse real-world scenarios, such as large-scale autonomous driving data (ACDC~\cite{Sakaridis2021}) or place recognition data (Tokyo~24/7~\cite{Torii.2015}), including challenging weather and illumination changes.

Training on this broader set of datasets helps keypoint detectors become more robust to real-world conditions, especially for challenging long-term and dynamic localization tasks, as we show in our evaluation (\cref{subsec:aachen_eval} and Sec.~6.1).

However, using only binary labels substantially weakens training supervision. To overcome this and address the non-differentiable nature of the keypoint selection process, we introduce a probabilistic formulation for keypoint selection via Reinforcement Learning (RL). Unlike previous RL-based methods~\cite{Tyszkiewicz.2020,Bhowmik.2020,Potje.2023} that still require depth or pose information, we propose to derive the reward exclusively from labeled image pairs.
Our key insight is to leverage the epipolar constraint, a fundamental concept in computer vision, that all \textit{true} keypoint matches in a positive pair must satisfy. This eliminates the need for pre-generated 3D models and further broadens the range of suitable training datasets.

To efficiently associate every keypoint location with its descriptor, we incorporate intermediate multi-scale information by leveraging hyper-column features from intermediate layers of the encoder, rather than relying solely on the final low-resolution output.
We further strengthen the descriptiveness of the descriptors by introducing a robust loss function explicitly designed for binary-labeled labeled image pairs.

We evaluate RIPE on MegaDepth~1500 and HPatches, achieving competitive results compared to the state-of-the-art.
On Aachen~Day-Night and Boreas we demonstrate its increased robustness to adverse weather and illumination changes thanks to our minimal supervision strategy.

In summary, the key contributions of our work include:

\begin{itemize}
    \item \textbf{Innovative Weakly-Supervised Training Framework:} Introduction of RIPE, a novel method based on Reinforcement Learning that trains keypoint detectors using only labeled image pairs, effectively removing the dependency on depth or pose information.
    \item \textbf{Improved Generalizability:} Ability to utilize diverse training datasets, improving keypoint detector performance in varying real-world conditions.
    \item \textbf{Epipolar Geometry-based reward:} Utilization of the epipolar constraint to derive rewards from labeled image pairs, ensuring that the optimization process adheres to fundamental principles of computer vision.
    \item \textbf{Multi-Scale Feature Representation:} Integration of multi-scale hyper-column features to enhance the association between keypoint locations and descriptors, leading to more informative and discriminative representations.
    \item \textbf{Robust Descriptor Loss:} Development of a robust loss function based on labeled image pairs, further strengthening the descriptiveness and reliability of the keypoint descriptors.
\end{itemize}

\section{Related Work}
\label{sec:related_work}

Evolving from classical hand-crafted approaches, current training-based methods for learning keypoint detection and description either rely on artificial augmentations or the availability of 3D information such as pose or depth, often derived from pre-generated 3D data. We systematically categorize current state-of-the-art methods based on their underlying training principles.

\PAR{Artificial augmentations}
Methods in this category generate training data by applying artificial augmentations to existing images, creating image pairs with known transformations. This can be achieved by using photometric and/or homographic data augmentation (SuperPoint~\cite{DeTone.2018}, DomainFeat~\cite{Xu.2024f5b}, SiLK~\cite{Gleize.2023}). As these are limited in their representation of illumination changes, style-transfer techniques were introduced to synthesize night images (DomainFeat~\cite{Xu.2024f5b}, Melekhov~\etal~\cite{Melekhov.2021}).

Additionally, different losses are used to ensure that feature maps are robust against domain changes. For instance, the triplet loss (DomainFeat~\cite{Xu.2024f5b}) enhances descriptor descriptiveness, while the minimization of margins for corresponding patches (HardNet~\cite{Mishchuk.2017}) or hard-negative sampling of local features (Melekhov~\etal~\cite{Melekhov.2021}) is also utilized.

\PAR{Pose or depth information}
Methods of the second category depend on known pose or depth information, typically pre-generated by Structure-from-Motion (SfM) techniques. Many methods \cite{Dusmanu.2019, Germain.2020, Wang.2020, Edstedt.2024,Wang2020} utilize the work of Li and Snavely \cite{Li2018}, leveraging the MegaDepth dataset (see \cref{fig:teaser}) for two example images), which contains photos of landmark collections, automatically annotated with pixel-wise depth information.
This dataset is constructed using \textsc{COLMAP} \cite{schoenberger2016sfm, schoenberger2016mvs} based on SIFT \cite{Lowe.2004}, to generate a 3D model along with pixel-wise depth information for each image.

Known depth information is used to calculate keypoint correspondences to train detection and description (Dusmanu~\etal~\cite{Dusmanu.2019}) or to calculate reward values (DISK~\cite{Tyszkiewicz.2020}). DeDoDe~\cite{Edstedt.2024} introduced the direct use of 3D tracks originating from reconstructed points in the 3D model as a supervision signal. ALIKE~\cite{Wang.2020} proposed to ground the learning of descriptors on relative pose information by introducing a differentiable matching layer and translating the relative poses into epipolar constraints. ALIKED~\cite{Wang.2023} additionally introduced attention-weighted local descriptors to include image-level spatial awareness into the descriptor, training on pixel-wise correspondences enriched with random homography transformations.
Recently, the dense (RoMa~\cite{Edstedt.2024c99}, Mast3r~\cite{Leroy.2024}, Dust3r~\cite{Wang.2023s18}) and semi-dense (S2DNet~\cite{Germain.2020}, LoFTR~\cite{Sun.2021}, Efficient LoFTR~\cite{Wang.2024}) matching methods moved into focus. These methods also use the available depth information from the pre-generated 3D models.

Tyszkiewicz~\etal~\cite{Tyszkiewicz.2020} (DISK) and Bhowmik~\etal~\cite{Bhowmik.2020} (Reinforced Feature Points) independently introduced Reinforcement Learning for learning keypoint detection and description to overcome the non-differentiability of keypoint detection. Bhowmik~\etal trained with a complete computer vision pipeline, treating the matching and pose estimation stages as non-differentiable black boxes, requiring known poses to calculate the reward value. In contrast, DISK computes rewards based on the number of correct feature matches (determined by the known relative position), allowing for precise calculation of matching probabilities. Potje~\etal~\cite{Potje.2023} (DEAL) extended the DISK approach by introducing an additional Warp Module, increasing robustness against non-rigid image deformations.

\PAR{SOTA Limitations} Ultimately, existing methods rely on known pixel-wise correspondences derived from artificial augmentations, depth information (measured or estimated with SfM) or relative pose. Consequently, these methods remain dependent on limited datasets or augmentation techniques. RL also remains underutilized, as depth (DISK~\cite{Tyszkiewicz.2020}), pose (Reinforced Feature Points~\cite{Bhowmik.2020}) or artificial augmentations (DEAL~\cite{Potje.2023}) are still required.

To this end, we introduce a more radical Reinforcement Learning approach, which allows us to train without known poses, without depth information, and without pixel-wise correspondences -- just using paired images.
This is comparable to semantic keypoint matching works like Rocco~\etal~\cite{Rocco2018}, but differs in the usage of RL, the inclusion of negative pairs and the enforcement of a valid epipolar geometry.

\section{Method}
\label{sec:method}

\begin{figure}[tb]
    \centering
    \includegraphics[width=1.0\linewidth]{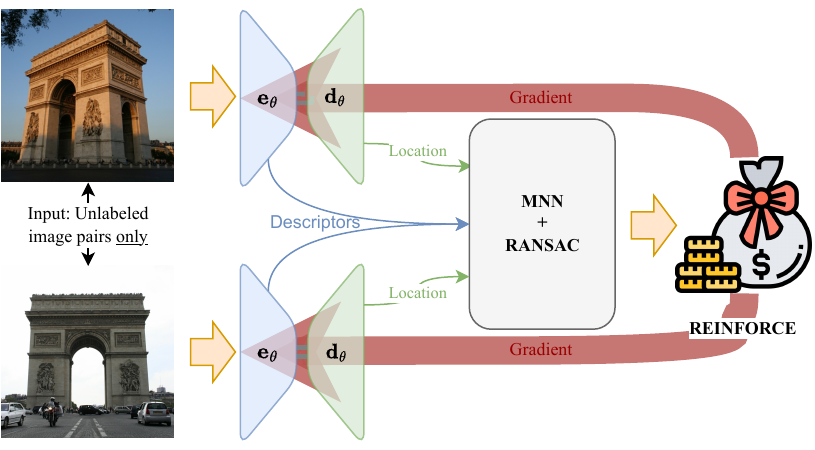}
    \vspace{-0.45cm}
    \caption{Overview of RIPE, our approach for learning keypoint detection and description from unlabeled image pairs using reinforcement learning. For an image pair, heatmaps are generated for probabilistic keypoint sampling, with descriptors derived from hyper-column features. These keypoints are matched and filtered via the fundamental matrix, with the number of matchable keypoints serving as the reward signal. This encourages the network to produce a large number of keypoints fulfilling the epipolar constraint for positive pairs and fewer keypoints for negative pairs.}
    \vspace{-0.3cm}
    \label{fig:overview}
\end{figure}

In the following, we present our approach to learning keypoint detection and description solely from unlabeled image pairs by using Reinforcement Learning (RL).
A visual overview of our method is provided in \cref{fig:overview}.
For each image in a given pair, a neural network generates a heatmap from which keypoint positions are sampled (\cref{subsec:dect}).
Each keypoint is then associated with a descriptor, extracted from the decoder using hyper-column features (\cref{subsec:desc}).
We process pairs of images and the resulting keypoints  are matched and filtered by estimating the fundamental matrix.
The final number of successfully matched keypoints is used as the reward signal, based on the label of the input pair.
Using Reinforcement Learning (\cref{eq:reinforce}), the reward encourages the network to generate a greater number of matchable keypoints -- consistent with the epipolar constraint -- and to produce fewer keypoints for negative pairs.
This formulation effectively leverages the feedback from both geometric consistency and image similarity to guide the learning process.

Consequently, the only required label is binary, indicating whether the paired images depict the same scene (\ie a sufficient number of 2D image keypoints correspond to projections of the same 3D point from the depicted scene) or not. 
Hence, we assume a dataset $\mathcal{D} = \{(I_\kappa, {I_\kappa}', \lambda_\kappa) \mid \kappa = 1, \ldots, N\}$ where each tuple contains two images and a binary label $\lambda_\kappa$ with
\begin{equation}
    \lambda_\kappa = 
    \begin{cases} 
    1, & \text{if } I_\kappa \text{ and } I_\kappa' \text{ show the same scene} \\
    -1, & \text{otherwise}
    \end{cases}   
\end{equation}
to indicate positive and negative pairs.

\subsection{Keypoint detection}
\label{subsec:dect}

\begin{figure}[tb]
    \centering
    \includegraphics[width=1.0\linewidth]{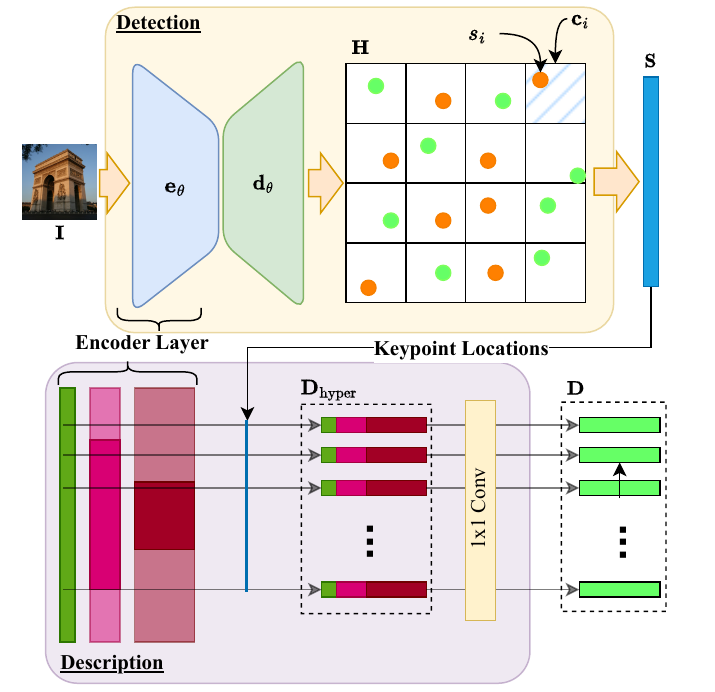}
    \vspace{-0.45cm}
    \caption{Keypoint detection (top) and description (bottom) for a single input image $\mathbf{I}$. The network outputs a logit heatmap $\mathbf{H}$ for each input image. Potential keypoint locations $\mathbf{S}$ are sampled from equally sized patches based on logit values, with rejected keypoints marked in orange and accepted ones in green. At the keypoint locations, descriptors $\mathbf{D}$ are generated from the intermediate encoder layers, linking each location to a descriptor integrating context at different scales from the intermediate layers.}
    \label{fig:dect+desc}
    \vspace{-0.3cm}
\end{figure}

Our proposed method RIPE starts with detecting possible keypoint positions in each image, employing an hourglass network composed of an encoder $\mathbf{e}_{\theta}(\cdot)$ and a decoder $\mathbf{d}_{\theta}(\cdot)$ connected via skip connections with learnable parameters $\theta$.
A graphical overview is given in the upper part of \cref{fig:dect+desc}.
For an input image $\mathbf{I} \in \mathbb{R}^{h \times w \times c}$, the network generates a detection heatmap $\mathbf{H} \in \mathbb{R}^{h \times w}$ with $\mathbf{H}=\mathbf{d}_\theta(\mathbf{e_\theta}(\mathbf{I}))$, indicating potential keypoint locations.
The heatmap $\mathbf{H}$ is divided into a regular grid, with each cell $\mathbf{c}_{i}$ having a size $m \times m$.
The total number of cells is denoted by $C = \left\lfloor \frac{h}{m} \right\rfloor \times \left\lfloor \frac{w}{m} \right\rfloor$, with $\left\lfloor \cdot \right\rfloor$ denoting the floor function.
Each cell $\mathbf{c}_{i} \in \mathbb{R}^{m \times m}$ holds the corresponding logit values.
The logit values in each cell constitute a categorical probability distribution from which exactly one keypoint location per cell is sampled, resulting in a keypoint position \( s_{i} \), with an initial probability \( \hat{p}_{i} \) and logit \( l_{i} \).
As a result, the network learns to define the keypoint locations by shaping the logit values accordingly.
Acknowledging that some image regions (such as overexposed sections, sky, etc.) may not be optimal for reliable keypoint detection, a sigmoid function is applied to the keypoint logit forming an acceptance indicator $a_{i}=\sigma(l_i)$ for each cell.
This enables the network to probabilistically discard unsuitable keypoints.
Consequently, the final probability \( p_{i} = \sigma(l_{i}) \cdot \hat{p}_{i} \) captures both the initial sampling chance and the likelihood of keypoint retention.

The keypoint detection results in a list $\mathbf{S}_\mathbf{I} \in \mathbb{R}_{2 \times C}$ of keypoint locations, acceptance indicators $\mathbf{a}_\mathbf{I} \in \mathbb{R}_{1 \times C}$, and associated probability values $\mathbf{p}_\mathbf{I} \in \mathbb{R}_{1 \times C}$ gathered from the individual cells of each keypoint in the image $\mathbf{I}$.
The total number of keypoints is equivalent to the number of cells $C$.

\subsection{Keypoint description}
\label{subsec:desc}

Once the keypoints are detected, the next step is to assign a descriptor to each location in order to enable the matching between images.
This process is visualized in the lower part of \cref{fig:dect+desc}.
Potje~\etal~\cite{Potje.2023} demonstrated that features derived from the encoder and bilinearly upsampled to the input resolution retain a high distinctiveness under small to moderate photometric and geometric changes.
However, this approach proved insufficient for our requirements, as it does not facilitate a reliable matching, as discussed in Sec.~6.2.
To enhance (with minimal computational overhead) the quality of the features created simply by upsampling the final feature layer,  we employ hyper-column features \cite{Hariharan.2015}, a concept also proposed by Germain~\etal~\cite{Germain.2019} and tested in diverse settings (\eg~Li~\etal~\cite{LiHyper2020}, Min~\etal~\cite{MinHyper2019})  to combine multi-scale information for increased descriptiveness.

In each layer $l$ of the encoder, feature vectors $\mathbf{e}_i^l$ at the detected keypoint locations $\mathbf{s}_i \in \mathbf{S}$ are bilinearly interpolated (please refer to \cref{fig:dect+desc} for a visualization). 
This results in a list $(\mathbf{e}_i^1, \mathbf{e}_i^2, \ldots \mathbf{e}_i^L)$ of feature vectors that encode information at different image scales. These feature vectors are then concatenated to form the intermediate hyper-column features $\mathbf{D}_\text{hyper} \in \mathbb{R}_{\hat{d} \times C}$, 
with $\hat{d}$ being the overall sum of channels for the encoder layers.
As the final dimensionality $\hat{d}$ is usually large (960 for a VGG-19), a final 1x1 convolution is applied to reduce the dimensionality to $d$, resulting in compact descriptors $\mathbf{D} \in \mathbb{R}_{d \times C}$.

For a single image $\mathbf{I}$, the detection and description results in: keypoint locations $\mathbf{S} \in \mathbb{R}_{2 \times C}$ (in image coordinates), acceptance indicators $\mathbf{a} \in \mathbb{R}_{1 \times C}$, selection probabilities $\mathbf{p} \in \mathbb{R}_{1 \times C}$, and the descriptor map $\mathbf{D} \in \mathbb{R}_{d \times C}$.
\textit{During training}, we repeat the entire process to obtain $\mathbf{S'}$, $\mathbf{a'}$, $\mathbf{p'}$, and $\mathbf{D'}$ for the second image $\mathbf{I}'$ of a pair.

\subsection{Reinforcement of matchable keypoints}
\label{subsec:dect_loss}

\begin{figure*}[htb]
    \centering
    \includegraphics[width=1.0\linewidth]{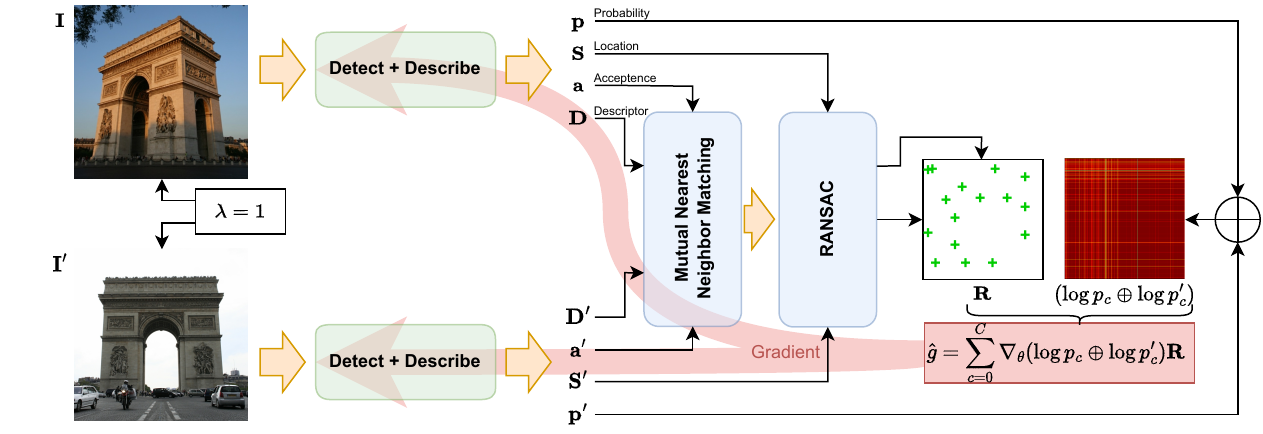}
    \vspace{-0.45cm}
    \label{fig:reinforce}
    \caption{Overview of the proposed Reinforcement Learning formulation for learning keypoint detection and description from unlabeled image pairs. A network/ agent generates probability distributions over potential keypoint locations. Actions (keypoint locations $\mathbf{S}$) are sampled from these distributions and associated with their respective probability $\mathbf{p}$ . The agent receives rewards based on the number of mutual nearest neighbors further filtered by estimating the fundamental matrix. This encourages the detection of matching keypoints in positive pairs ($\lambda=1$) while penalizing incorrect detections in negative pairs ($\lambda=-1$). Using REINFORCE~\cite{Williams.1992}, gradients are derived and utilized to update the network parameters accordingly.}
    \vspace{-0.3cm}
\end{figure*}

To address the discrete nature of the keypoint detection process and to allow training from unlabeled image pairs only, we formulate it as a reinforcement problem.
Starting from a brief general introduction, we develop the methodical foundation of RIPE.

The policy is defined as a probability distribution over actions $\mathcal{A}$, conditioned on the current state $\mathcal{S}$ and parameterized by $\theta$ with
\begin{equation}
    \pi_\theta(\mathcal{S})=\mathbb{P}[\mathcal{A}|\mathcal{S}, \theta].
\end{equation}
This constitutes a probability distribution, from which an action $\mathcal{A}$ is sampled.
Based on the sampled action, the agent receives a reward signal that indicates a good or bad action.
The learning objective is then formulated as maximizing the expected cumulative reward over a trajectory $\tau$ (a sequence of state, action, reward tuples) scaled by the reward $R$:
\begin{equation}
    \max_{\theta} J(\theta) = \mathbb{E}_{x \sim \pi_\theta}[R(\tau)], 
\end{equation}
However, directly calculating the derivative of $J$ is not feasible, as it would require differentiating through all possible trajectories and the state distribution.
Nonetheless, REINFORCE~\cite{Williams.1992} provides an approximation for the derivative:
\begin{equation}
    \nabla_\theta J(\theta) \approx \hat{g} =\sum_{t=0} \nabla_\theta \log \pi_\theta(a_t | s_t) R(\tau).
\label{eq:reinforce}
\end{equation}
In RIPE, the encoder-decoder network acts as a trainable policy, with the input image $\mathbf{I}$ representing the state.
This leads to the following formulation for the policy:
\begin{equation}
\begin{split}
    \pi_\theta(\mathbf{s}) & = d_\theta(e_\theta(\mathbf{I})) = \mathbf{p} \\
    & = \Bigl[\mathbb{P}_1[a_1|I,\theta], \mathbb{P}_2[a_2|I,\theta], \dots, \mathbb{P}_c[a_c|I,\theta]) \Bigr],
\end{split}
\end{equation}
where $\mathbf{p}$ is a list of distributions for each cell $\mathbf{c}$ in the heatmap $\mathbf{H}$.
As described in \cref{subsec:dect}, keypoint locations are sampled from these distributions (\ie the keypoint localization corresponds to an action).

Working with image pairs, we introduce a second list of probabilities $\mathbf{p'}$  and compute the joint probability across all combinations of cells. This allows us to approximate the gradient, as given by:

\begin{equation}
    \hat{g} =\sum_{c=0}^C \nabla_\theta (\log p_c \oplus \log p'_c) R,
\label{eq:grad}
\end{equation}
with $\oplus$ denoting the outer sum\footnote{With $x \in \mathcal{R}_{m \times 1}$ und $y \in \mathcal{R}_{m \times 1}$ the outer sum gets calculated as $x \oplus y \in \mathcal{R}_{m \times n}$ with $(x \oplus y)_{i,j}=x_i + y_j$}.

\begin{figure*}[th]
    \centering

    \begin{subfigure}[b]{\textwidth}
            \includegraphics[width=0.33\textwidth]{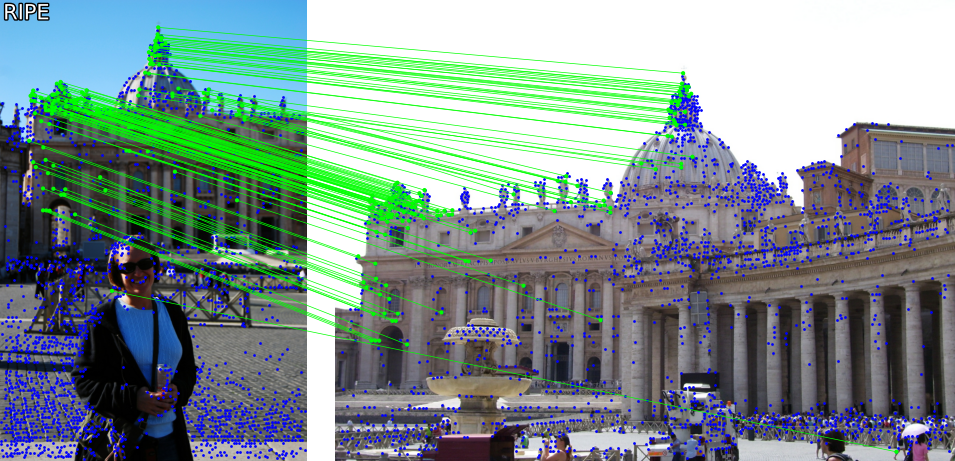}
            \includegraphics[width=0.33\textwidth]{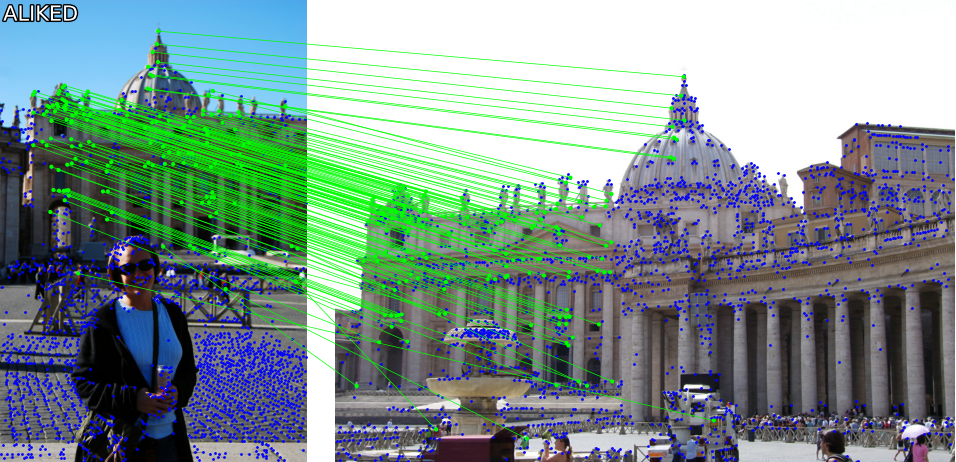}
            \includegraphics[width=0.33\textwidth]{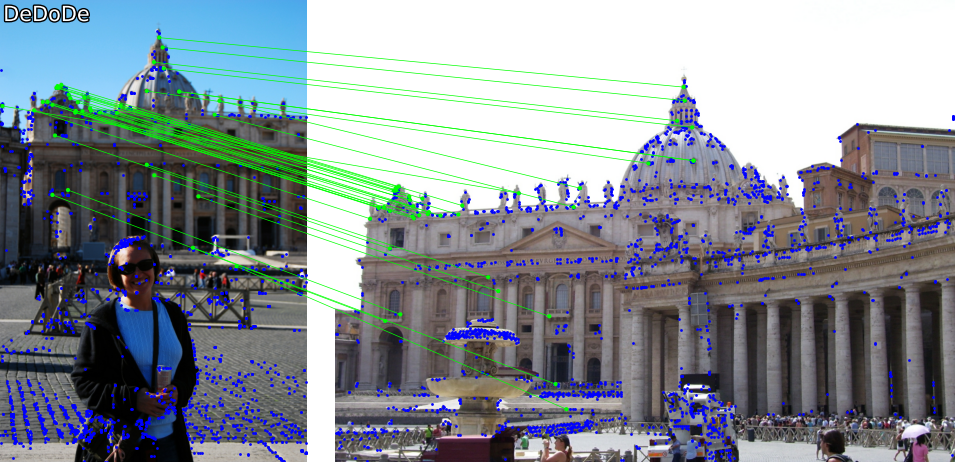}
    \end{subfigure}
\caption{Example results for RIPE (ours), ALIKED~\cite{Zhao.2023dvq}, and DeDoDe~\cite{Edstedt.2024} from the MegaDepth~1500 benchmark. RIPE demonstrates its ability to effectively utilize the unlabeled paired images to disregard irrelevant regions and identify highly discriminative keypoints. DeDoDe tends to cluster its keypoints in specific areas, whereas the keypoints of ALIKED and RIPE are more evenly distributed across the image.}
\label{fig:qualitative_result}
\vspace{-0.5cm}
\end{figure*}

The only ingredient missing is the reward $R$.
As motivated in our introductory example, the goal is to produce many matching keypoints for positive pairs and few for a negative pairs.
Consequently, the network should be rewarded for detecting keypoints in positive image pairs that are both matchable and pass the geometric filtering based on the epipolar constraint. 
Conversely, a negative reward (penalty) should be applied when the network incorrectly detects keypoints that appear matchable and filterable, yet originate from different scenes.
The reward is computed using mutual nearest-neighbor estimation and RANSAC filtering, which, despite being non-differentiable, are used here solely for the reward calculation, and thus do not require gradient computation.

RIPE identifies mutual nearest neighbors between the keypoint descriptors in $\mathbf{D}$ and $\mathbf{D'}$ by computing their L2-distance, yielding a list of index pairs $\mathbf{C}$.
This list is again filtered by estimating the fundamental matrix $\mathbf{F}$ using the 8-point algorithm in combination with RANSAC.
The resulting mask $\mathbf{M}$ indicates all mutual nearest neighbors that can be explained by a common epipolar geometry.

The reward matrix $\mathbf{R} \in \mathbb{R}^{C \times C}$ with the individual reward values $r_{i,j} \in \mathbf{R}$ is created by
\begin{equation}
r_{p,q} = 
\begin{dcases} 
\sign(\lambda_\kappa) \rho, & (p, q) \in \mathbf{C} \text{ and } \mathbf{M}_{p,q} \text{ is True} \\ 
0, & \text{otherwise}
\end{dcases}
\label{eq:reward}
\end{equation}
with the reward $\rho$ and the signum function $\sign(\cdot)$.
Simply inverting the reward for negative pairs produces in a symmetric training signal, which we found to be beneficial during training.

The resulting gradients of \cref{eq:grad} adjust the network parameters $\theta$ to increase the likelihood of selecting keypoints (i.e., actions taken) that result in a positive reward. 
Since the expectation is approximated by the mean, a substantial number of samples is required for a reliable estimate. 
In our approach, the sampling of $C$ keypoint locations for every image acts as an equivalent of sampling multiple trajectories (as in classical reinforcement learning), and allows a sufficient approximation of the expectation.

\subsection{Descriptor loss}
\label{subsec:desc_loss}

To enhance the descriptiveness and robustness of the descriptors, we integrate a second loss
\begin{equation}
    L_{\text{desc}}=
    \begin{dcases}
    \frac{1}{N}\sum_{n=1}^{N}\max(0,\mu+\delta_{+}^n-\delta_{h}^n), & \lambda_\kappa=1 \\
    \frac{1}{N}\sum_{n=1}^{N}\max(0,\mu - \delta_{+}^n), & \text{otherwise} 
    \end{dcases}
\label{eq:desc_loss}
\end{equation}
where $N$ is the number of inliers after the RANSAC filtering, $\mu$ is a positive threshold, and $\delta$ the L2-distance between two descriptors.
This loss behaves differently depending on the type of image pair, with $\delta_{+}^n$ being the L2-distance between two keypoint descriptors, which are matched \textit{and} were validated (i.e., identified as inliers) by geometric filtering, while $\delta_{h}^n$ is the distance to the second closest neighbor.
In the positive case, this loss forces the descriptors of putative matches (putative, as we have no ground truth verification for belonging to the same 3D point) to become more similar while repelling all other descriptors.
In contrast, in the negative case, where no matches should exist, the loss drives the descriptors of incorrectly matched keypoints further apart, thus increasing robustness by minimizing false positives.

\subsection{Final loss}
\label{subsec:final_losss}

We define the final loss for the descriptor as $\mathcal{L}_{\text{desc}}=-\mathbb{E}_{\mathbf{K}}[\mathbf{R}]$.
Inspired by the work of Potje~\etal~\cite{Potje.2023}, we add a regularization term
\begin{equation}
\mathcal{L}_{\text{low}}=-\sum_p \sum_q \log p_{i} \cdot \epsilon,
\label{eq:reg}
\end{equation}
with $k$ being a small negative constant to prevent the network from generating low probability keypoints.
The final loss is calculated by 
\begin{equation}
    \mathcal{L} = \mathcal{L}_{\text{dect}} + \mathcal{L}_{\text{low}} + \psi \mathcal{L}_{\text{desc}},
\label{eq:final_loss}
\end{equation}
where $\psi$ is a small positive constant, to balance detection and description loss.

By framing keypoint detection as an RL problem, RIPE can backpropagate the final loss to optimize both keypoint locations and descriptors, effectively tackling the challenge of learning without relying on extensive labeled datasets .
The use of RL and formulating the detection process as a probabilistic decision also overcomes the inherent discreteness of the keypoint selection, enabling direct optimization.
With this approach, we demonstrate how keypoints can be learned effectively using only unlabeled image pairs -- mirroring the way humans would perform the task.
\section{Experiments}
\label{sec:experiments}

\subsection{Implementation details}

We used the \textsc{torchvision} implementation of VGG-19 with strides of $[1,2,4,8]$ and pretrained on ImageNet as our encoder.
The feature maps on each scale have $[64,128,256,512]$ channels, respectively, resulting in $d'=960$ (see \cref{subsec:desc}), reduced to $d=256$, by the 1x1 convolutional layer.
For the decoder, we follow DeDoDe~\cite{Edstedt.2024} and use the depthwise convolutional refiners  proposed by Edstedt~\etal in DKM~\cite{Edstedt2022}, with 8 blocks per scale and internal dimension $[64,128,256,512]$.
We use the GPU-accelerated mutual nearest neighbor implementation from the Kornia library~\cite{eriba2019kornia} and the fundamental matrix estimation from PoseLib~\cite{PoseLib}.
The network is trained with AdamW~\cite{Loshchilov2017DecoupledWD} with a learning rate starting from 0.001 and linearly decaying to 1e-6.
We train for 80,000 steps with a batch size of 6 and a gradient accumulation over 4 batches, taking three days on a single A100 GPU.
Input images are normalized, resized and padded to meet the desired input resolution, without altering the aspect ratio.
In addition, we set the hyperparameters according to \cref{tab:hp}.
To facilitate the initial stages of training, we linearly increase the influence of $\epsilon$ during the first third of the training process.

\begin{table}
\small
\centering
\begin{tabular}{ccl}
\toprule
\multicolumn{1}{l}{Variable} & \multicolumn{1}{l}{Value} & \multicolumn{1}{c}{Purpose} \\
\midrule
$h, w$ & 560 & image size \\
$m$ & 8 & cell size \\
$\rho$ & 1.0 & reward \cref{eq:reward}\\
$\epsilon$ & -7e-8 & constant regularization \cref{eq:reg} \\
$\psi$ & 5.0 & descriptor loss weight \cref{eq:final_loss}\\
\bottomrule
\end{tabular}
\caption{Overview of our hyperparameter configuration.}
\label{tab:hp}
\vspace{-0.75cm}
\end{table}

\begin{table*}[ht]
\small
\centering
\setlength{\tabcolsep}{7.2pt}
\begin{tabular}{@{}llccccccc@{}}
\toprule
\multirow{2}{*}{Category}     & \multicolumn{1}{c}{\multirow{2}{*}{Method}}  & \multicolumn{3}{c}{MegaDepth 1500}                    & \multicolumn{1}{l}{\multirow{2}{*}{Label}} & \multicolumn{3}{c}{HPatches}                           \\ \cmidrule(lr){3-5} \cmidrule(l){7-9} 
                              & \multicolumn{1}{c}{}                         & AUC$@5^{\circ}$ & AUC$@10^{\circ}$ & AUC$@20^{\circ}$ & \multicolumn{1}{l}{}                       & AUC$@1\text{px}$ & AUC$@3\text{px}$ & AUC$@5\text{px}$ \\ \midrule
\multirow{3}{*}{(Semi-)Dense} & Eff. LoFTR\cite{Wang.2024}\tiny{CVPR'24}     & 64.63           & 78.03            & 87.21            & D                                        & 42.38           & 63.26           & 72.75           \\
                              & Mast3r\cite{Leroy.2024}\tiny{ECCV'24}        & 22.84           & 35.29            & 47.62            & D                                        & 29.63           & 54.04           & 66.86           \\
                              & RoMa\cite{Edstedt.2024c99}\tiny{CVPR'24}     & 68.28           & 80.53            & 88.77            & D                                          & 38.38           & 63.75           & 73.37           \\ \midrule
\multirow{7}{*}{Sparse}       & SIFT\cite{Lowe.2004}\tiny{IJCV'04}           & 37.29           & 50.61            & 61.89            & ---                                        & 29.98           & 53.93           & 65.01           \\
                              & SuperPoint\cite{DeTone.2018}\tiny{CVPRW'18}  & 49.5            & 62.51            & 72.12            & H                                          & 36.61           & 56.92           & 67.28           \\
%                             & SuperPoint+SuperGlue                         & 55.21           & 70.92            & 82.58            & \multicolumn{1}{l}{}                       & 39.7            & 61.99           & 73.06           \\
                              & DISK\cite{Tyszkiewicz.2020}\tiny{NeurIps'20} & 51.83           & 64.49            & 74.3             & P/D                                        & 37.54           & 57.68           & 68.08           \\
                              & ALIKED\cite{Zhao.2023dvq}\tiny{TIM'23}       & \textbf{56.71}  & \textbf{69.86}   & \textbf{79.54}   & P+H                                        & 31.95           & 57.65           & \underline{69.35} \\
                              & SiLK\cite{Gleize.2023}\tiny{ICCV'23}         & 29.83           & 40.59            & 49.98            & H                                          & \underline{39.05} & 57.96           & 67.64           \\
                              & DeDoDe-B\cite{Edstedt.2024}\tiny{3DV'24}     & 55.02           & 67.99            & 77.31            & P                                          & \textbf{39.88}  & \textbf{60.59}  & \textbf{70.36}  \\ \cmidrule(l){2-9} 
                              &\textbf{RIPE}                                 & \underline{55.11} & \underline{68.34} & \underline{78.03} & \textbf{Pair}                                   & 37.93           & \underline{58.93}           & 69.20           \\ \bottomrule
\end{tabular}
\vspace{-0.35cm}
\caption{Results for the evaluation of relative pose estimation on MegaDepth-1500 (left) and homography estimation on HPatches (right).
We specify the training data for each approach, with D indicating depth, P position and H homography/ artificial augmentations. 
Crucially, our method is the only one that requires only unlabeled image pairs, yet achieves competitive performance to the sparse SotA methods. 
The \textbf{best} and \underline{second-best} performances for the sparse methods are highlighted.}
\label{tab:megadepth-1500}
\vspace{-0.5cm}
\end{table*}

\subsection{Training data}

We use the same Megadepth subset as \cite{Tyszkiewicz.2020}, reducing the dataset to image pairs which show a sufficient number of covisible 3D points.
However, our innovative training regime enables the use of an extended data basis.
In \cref{subsec:aachen_eval} and Sec.~6.1 we therefore demonstrate how this can improve the robustness of our method. 

\subsection{Inference}
During inference, we pass a single image through the network and sample the top \textit{K} most likely keypoints and their positions from $\mathbf{H}$, accompanied with non-maximum suppression in a 3x3 window.
Each keypoint is associated with its hyper-column descriptor, as described in \cref{subsec:desc}.
The mean inference time (measure for the setting of \cref{subsec:rel_pose_eval} was \SI{0.47}{\second} for RIPE (for comparison: DeDoDe \SI{0.45}{\second}, ALIKED \SI{0.08}{\second}, DISK \SI{0.17}{\second}).

\subsection{Relative pose estimation}
\label{subsec:rel_pose_eval}

\PAR{Dataset} We evaluated the relative pose estimation performance using the MegaDepth-1500 subset.
It contains two (\textit{Brandenburger Tor} and \textit{St. Peters Square}) out of the 196 scenes from the original MegaDepth dataset and was introduced in LoFTR~\cite{Sun.2021}.
The main challenges are large viewpoint, illumination changes and repetitive patterns.
Following recent evaluation protocols (\cite{Wang.2024}, \cite{lindenberger_2023_lightglue}), we resized the longer side of the input images to 1600 for the dense methods and to 1200 for the (semi-)sparse methods.
To assess the quality of the extracted keypoints, we use mutual-nearest-neighbor matching for all sparse methods.

\PAR{Baselines} We compare RIPE against SotA methods for sparse keypoint detection and description.
Due to their recent success, we also integrate dense (RoMA~\cite{Edstedt.2024c99}, Mast3r~\cite{Leroy.2024}) and semi-dense matching methods for comparison (Efficient LoFTR~\cite{Wang.2024}).
For the sparse methods, we use mutual-nearest-neighbor matching to establish correspondences.

\PAR{Metrices} Building on preceding approaches, the accuracy of matches is assessed by evaluating the relative poses they yield. The pose error is characterized as the greatest of the angular discrepancies in both rotation and translation.
We provide the Area Under the Curve (AUC) of the pose error at the thresholds of \ang{5}, \ang{10}, and \ang{20}.
We used the glue-factory library, kindly provided by the authors of GlueStick~\cite{pautrat_suarez_2023_gluestick} and LightGlue~\cite{lindenberger_2023_lightglue}.
We use the poselib \cite{PoseLib} for the robust pose estimation and select the top 2048 keypoints.

\PAR{Results} The left half of \cref{tab:megadepth-1500} and demonstrate that our method, RIPE, ranks a close second to the current state-of-the-art sparse method, ALIKED, while not requiring images with known poses during training and leveraging a significantly weaker training signal. 
Remarkably, RIPE even outperforms DeDoDe, despite the latter employing two distinct networks for detection and description, which nearly doubles the number of required parameters.
Furthermore, our method is the first to achieve this performance without relying on artificial homographies (as seen in methods like SuperPoint and SiLK) or requiring pose or depth information from a pre-registered 3D model.
Qualitative results can be found in \cref{fig:qualitative_result} and in the supplementary (Fig. 6).

\subsection{Homography Estimation}

\PAR{Dataset}
We evaluate on the HPatches dataset~\cite{hpatches_2017_cvpr}, which contains sequences of planar scenes, taken with viewpoint or illumination changes.
We resized the input smaller side of the input images to 480~pixels.

\PAR{Metrices} To assess the quality of the homography estimation we calculate the mean reprojection error of the corner points and report the AUC for the thresholds of 1, 3 and 5 pixels.
For all methods, we use mutual-nearest-neighbor matching and the implementation of poselib \cite{PoseLib} for the robust homography estimation and restricted the number of keypoint to 1024.

\PAR{Results} As illustrated on the right side of \cref{tab:megadepth-1500}, RIPE once again performs on par with the state-of-the-art methods SiLK and DeDoDe, despite not utilizing artificial homographies like SiLK or pose information as employed by DeDoDe. 
The strong performance of SiLK highlights the advantages of its training regimen, which relies solely on artificial augmentations and therefore closely resembles the testing data. 
Conversely, the impressive performance of ALIKED diminishes on the HPatches benchmark, where it only narrowly surpasses SIFT.

\subsection{Outdoor localization day-night}
\label{subsec:aachen_eval}

\PAR{Dataset} To evaluate our approach in the context of visual localization (\ie the estimation of the 6-DoF pose for a query image relative to a 3D~scene model), we use the Aachen~v1.1~\cite{Zhang2020ARXIV,Sattler2018CVPR,Sattler2012BMVC} dataset.
This dataset is especially challenging because of its large viewpoint and illumination (day-night) changes.
We used the HLoc localization framework~\cite{sarlin2019coarse} for the evaluation and first triangulate a 3D~model from the 6,697 reference images.
For each of the 1015 (824 daytime, 191 nighttime) queries we retrieve 50 images using NetVLAD \cite{Arandjelović.2016} and match them.
To evaluate the influence of additional training data, we replaced 20\% of the training samples with images from the Tokyo~24/7~\etal~\cite{Torii.2015} query dataset, originally intended for place recognition.
This dataset contains images from 125 distinct locations.
%At each location, images had been taken in three different viewing directions, during day, dusk and night, resulting in 1125 images in total.
Images were captured at each location from three different viewing directions, across day, dusk, and night, resulting in a total of 1,125 images.
The images are paired based on their geo-position only.
No 3D positions or depth maps are available.

\PAR{Metrics} The camera pose is estimated with a Perspective-n-Point solver in conjunction with RANSAC and the AUC is reported for thresholds 0.25\si{\m}/\ang2, 0.5\si{\m}/\ang{5} and 1.0\si{m}/\ang{10}.

\PAR{Results} Our results in \cref{tab:aachen} highlight the importance of diverse training data for the robustness of trained keypoint extractors. 
As ALIKED was not only trained on MegaDepth, but also on synthetic and training images from the Aachen\cite{Sattler2018CVPR} dataset, it shows strong performance for the localization of night-time queries.
The results from DeDeDo demonstrate the influence of the limited training data (MegaDepth only).
RIPE clearly outperforms DeDoDe, even if only trained on MegaDepth.
When day-to-nighttime images are added to training data, the margin further increases.
This improvement is made possible by our innovative training regime, which facilitates the addition of diverse training data.

Additional experiments presented in the supplementary material (Sec. 6.1) further support these findings, with results obtained from the Boreas~\cite{Burnett.2023} dataset, which encompasses challenging weather conditions.

\begin{table}[t]
\small
\centering
\setlength{\tabcolsep}{3pt}
\begin{tabular}{@{}lcccccc@{}}
\toprule
\multicolumn{1}{c}{\multirow{2}{*}{Method}} & \multicolumn{3}{c}{Day}  & \multicolumn{3}{c}{Night} \\ \cmidrule(l){2-4} \cmidrule(l){5-7} 
\multicolumn{1}{c}{}                                 & .25m/\ang{2} & .5m/\ang{5} & 5m/\ang{10} & .25m/\ang{2}  & .5m/\ang{5} & 5m/\ang{10} \\ \midrule
ALIKED                                               & \textbf{87.3}    & \textbf{93.9}   & \textbf{97.3}  & \textbf{73.3}     & \textbf{88.0}   & \textbf{96.9}  \\
DeDoDe                                               & 82.2    & 89.0   & 92.6  & 47.1     & 56.5   & 64.4  \\
SIFT                                                 & 82.5    & 88.5   & 91.9  & 30.9     & 38.2   & 46.1  \\ \midrule
RIPE                             & \underline{81.6}    & \underline{89.2}   & \underline{93.1}  & 52.9     & 67.5   & 79.1  \\
$\downarrow$\scriptsize{ + Tokyo} & \gss{+0.0} & \rss{-0.2} & \rss{-0.7} & \gss{+7.3} & \gss{+5.3} & \gss{+4.7}   \\
RIPE                          & \underline{81.6}    & 89.0   & 92.4  & \underline{60.2}     & \underline{72.8}   & \underline{83.8}  \\ \bottomrule
\end{tabular}
\vspace{-0.35cm}
\caption{Outdoor visual localization on the Aachen Day-Night v1.1. The results emphasize the significance of available training data: ALIKED outperforms DeDoDe by incorporating images from the Aachen dataset. Additionally, RIPE demonstrates substantial improvements by including day-night training pairs from Tokyo 24/7. The \textbf{best} and \underline{second-best} performances are highlighted.}
\label{tab:aachen}
\vspace{-0.25cm}
\end{table}

\subsection{Evaluation dataset Composition}

\begin{table}[ht]
\centering
\small
\setlength{\tabcolsep}{1.1pt}
\begin{tabular}{@{}cccccccc@{}}
\toprule
\multirow{2}{*}{MegaDepth} & \multirow{2}{*}{Tokyo} & \multicolumn{3}{c}{Day}  & \multicolumn{3}{c}{Night} \\ \cmidrule(l){3-8} 
                           &                             & 0.25m/2 & 0.5m/5 & 5m/10 & 0.25m/2  & 0.5m/5 & 5m/10 \\ \midrule
1.0                        & 0.0                         & 81.6    & \textbf{89.2}   & 93.1  & 52.9     & 67.5   & 79.1  \\
0.9                        & 0.1                         & 81.3    & 87.5   & \textbf{93.2}  & 57.1     & 68.1   & 82.7  \\
0.8                        & 0.2                         & \textbf{81.6}    & 89.0     & 92.4  & \textbf{60.2}     & \textbf{72.8}   & \textbf{83.8}  \\
0.7                        & 0.3                         & 80.6    & 88.1   & 93    & 56.5     & 71.7   & 82.2  \\
0.6                        & 0.4                         & 79.4    & 86.0   & 92.1  & 58.1     & 70.7   & 85.3  \\
0.5                        & 0.5                         & 78.0    & 84.2   & 89.4  & 56.5     & 68.6   & 83.2  \\
0.4                        & 0.6                         & 71.8    & 81.2   & 87    & 54.5     & 70.2   & 83.8  \\ \bottomrule
\end{tabular}
\vspace{-0.25cm}
\caption{Evaluation on how training data from the Tokyo~24/7 dataset improves the ability of RIPE to handle day to night illumination changes. The \textbf{best} performances are highlighted.}
\label{tab:aachen_tokyo}
\vspace{-0.6cm}
\end{table}
We further investigated which data mix between MegaDepth and Tokyo~24/7 yields the best improvement on the Aachen~Day-Night benchmark.
We trained RIPE with different compositions, as presented in \cref{tab:aachen_tokyo}, and found that 80\% MegaDepth data and 20\% Tokyo data were the most advantageous. 
A further increase in the proportion of Tokyo data begins to degrade the results, likely due to the lack of viewpoint variability in this dataset, as the images primarily differ in illumination. 
Consequently, RIPE struggles to learn to cope with the strong viewpoint variations that accompany the day-to-night changes in the Aachen benchmark.

Additional experiments regarding the influence of our hyperparameters can be found in the supplementary (Sec. 6.2).

\section{Conclusion}
\label{sec:conclusion}

We present a fundamentally new approach to learning keypoint detection and description.
By integrating Reinforcement Learning with essential computer vision principles, we successfully train RIPE using only unlabeled image pairs, expanding the pool of available training data.
This eliminates the constraints imposed by traditional approaches that depend on precise geometric annotations, making our method scalable and adaptable to diverse real-world scenarios.

Despite leveraging a significantly weaker training signal, RIPE achieves performance on par with state-of-the-art sparse keypoint extractors. 
Furthermore, our results demonstrate that RIPE effectively benefits from the inclusion of diverse training data, improving its generalization capabilities and robustness to challenging conditions. This highlights the potential of our approach to redefine keypoint learning, enabling broader applicability across various domains.

\small \PAR{Acknowledgments} We thank our colleague, Wieland Morgenstern, for his valuable feedback on the manuscript. This work was partly funded by the Investitionsbank Berlin (IBB) (BerDiBa, grant no. 10185426) and the German Federal Ministry for Economic Affairs and Climate Action (DeepTrain, grant no. 19S23005D).

{
    \small
    \bibliographystyle{ieeenat_fullname}
    \bibliography{main}
}

\clearpage
\setcounter{page}{1}
\maketitlesupplementary

\section{Additional experiments}

\subsection{Outdoor localization adversarial weather}
\label{subsec:boreas_eval}

Our training approach enables the straightforward integration of additional training data, allowing us to effectively adapt to challenging conditions.

\PAR{Dataset} For this experiment, we utilize the HLoc localization framework~\cite{sarlin2019coarse} with data from the Boreas dataset~\cite{Burnett.2023}, which includes high-resolution images, lidar, and radar data, accurately localized using GPS in an autonomous driving context.
The dataset features multiple acquisitions of the same route throughout the year, introducing varied weather and lighting challenges.
We select one sequence with favorable weather as our reference and four others (dark, late autumn, heavy snowfall, and rain) as query sequences. 
Please refer to \cref{tab:boreas_mapping} for a Boreas IDs to sequence name mapping.

\begin{table}[htbp!]
\centering
\small
\begin{tabular}{@{}cc@{}}
\toprule
Name           & Boreas ID               \\ \midrule
Reference      & boreas-2021-05-06-13-19 \\
Dark           & boreas-2020-11-26-13-58 \\
Late Autumn    & boreas-2020-12-18-13-44 \\
Heavy Snowfall & boreas-2021-01-26-11-22 \\
Rain           & boreas-2021-04-29-15-55 \\ \bottomrule
\end{tabular}
\vspace{-0.2cm}
\label{tab:boreas_mapping}
\caption{Mapping from the sequence names/ conditions to the actual identifiers in the Boreas dataset.}
\vspace{-0.35cm}
\end{table}

To manage the high sampling rate, we subsample the reference sequence to approximately 8k images, forming matching pairs with the next eight images for keypoint extraction and matching. 
We then triangulate our reference model using the geolocalized positions. Each query sequence is subsampled to about 3k images, and we retrieve 20 candidates per query image using NetVLAD~\cite{Arandjelović.2016} for localization.

To assess the influence of training data, we replace 10\% with images from the ACDC dataset~\cite{Sakaridis2021}.
This dataset provides 400 training images for adverse conditions (snow, rain, night, fog), each paired with a corresponding reference image from optimal conditions through geo-positioning.
This enables RIPE to learn keypoint detection across varying weather scenarios.

\PAR{Metrics} Each query camera pose gets estimated with a Perspective-n-Point solver in conjunction with RANSAC.
We report the AUC of the pose error for thresholds of 3\si{\centi\m}/\ang{3}, 5\si{\centi\m}/\ang{5} and 25\si{\centi\m}/\ang{2}.

\PAR{Results} The results in \cref{tab:boreas} illustrate the challenges posed by adverse weather conditions, leading to low performance under tighter thresholds. However, RIPE demonstrates competitive performance with state-of-the-art methods.
Furthermore, incorporating training data from ACDC, which features images under similar conditions to Boreas, enhances RIPE's results.
This underscores the significance of flexible training regimes that facilitate the integration of diverse datasets.

\subsection{Ablations}
\label{subsec:ablations}

\begin{table}[ht!]
\centering
\small
\setlength{\tabcolsep}{7.3pt}
\begin{tabular}{@{}lcccccc@{}}
\toprule
$\psi$      & 0.0 & 0.005 & 0.05  & 0.5   & 5              & 50    \\ \midrule
AUC@\ang{5} & --  & --    & 61.94 & 60.46 & \textbf{63.48} & 60.65 \\ \bottomrule
\end{tabular}
\vspace{-0.25cm}
\caption{Influence of the descriptor loss weight $\psi$ (\cref{eq:final_loss}) on the AUC@\ang{5} for the IMC2020 dataset. If no result is presented, our method failed to train successfully.}
\vspace{-0.35cm}
\label{tab:desc_loss}
\end{table}
We used a small subset of the 2020 Image Matching Challenge (IMC)~\cite{Jin.20211y} as our validation dataset during training to optimize our hyperparameters. We halted the training after 40,000 steps and report the final AUC@\ang{5} for relative pose estimation to assess the influence of our design choices. 

\begin{table}[h!]
\centering
\small
\setlength{\tabcolsep}{10.6pt}
\begin{tabular}{@{}lccccc@{}}
\toprule
$\epsilon$        & 0    & -7e-5 & -7e-6 & -7e-7 & -7e-8 \\
AUC@\ang{5}       & 61.0 & --    & --    & \textbf{63.48} & 57.45 \\ \bottomrule
\end{tabular}
\vspace{-0.25cm}
\caption{Influence of the descriptor loss weight $\epsilon$ (\cref{eq:reg}) on the AUC@\ang{5} for the IMC2020 dataset. If no result is presented, our method failed to train successfully.}
\label{tab:epsilon}
\vspace{-0.35cm}
\end{table}
\cref{tab:desc_loss} illustrates the impact of $\psi$, which weights the contribution of our descriptor loss (see \cref{subsec:desc_loss}) to the final loss (\cref{eq:final_loss}). RIPE fails to train effectively without our proposed descriptor loss, as the descriptors receive no direct training signal in its absence. This leads to poor matching during training, inhibiting the Reinforcement Learning process. Conversely, an insufficient influence of the descriptors is also detrimental.

\cref{tab:epsilon} shows the influence of the regularization parameter $\epsilon$ (\cref{eq:reg}).
RIPE still trains successfully without this regularization if $\epsilon=0$, but fails for too large values, as this discourages the network from detecting keypoints at all, resulting in a failed training.

We also experimented with removing our hyper-column descriptor extraction and replaced it with a bilinear upsampling of the final encoder layer.
With this configuration RIPE fails to train, as the descriptors are not discriminative enough.

\begin{table*}[t]
\small
\centering
\setlength{\tabcolsep}{8.2pt}
\begin{tabular}{@{}lcccccccccccc@{}}
\toprule
Method       & \multicolumn{3}{c}{Dark} & \multicolumn{3}{c}{Late Autumn} & \multicolumn{3}{c}{Heavy Snowfall}  & \multicolumn{3}{c}{Rain} \\ \cmidrule(r){1-1} \cmidrule(r){2-4} \cmidrule(r){5-7} \cmidrule(r){8-10} \cmidrule(r){11-13}
ALIKED\cite{Zhao.2023dvq}\tiny{TIM'23}       & 9.75   & \textbf{30.35}  & \textbf{89.6}   & 3.91     & \underline{12.54}     & \textbf{91.47}    & 2.37      & 9.77       & \textbf{72.43}      & 9.84   & 27.47  & \textbf{95.1}   \\
DeDoDe\cite{Edstedt.2024}\tiny{3DV'24}       & \textbf{10.64}  & \underline{30.23}  & 86.96  & 3.76     & 11.84     & 88.91    & 1.84      & 9.31       & 54.07      & \textbf{15.58}  & \textbf{35.68}  & 94.64  \\
DISK\cite{Tyszkiewicz.2020}\tiny{NeurIps'20}         & 8.97   & 27.68  & 87.44  & 3.26     & 11.19     & 89.56    & 2.34      & 10.16      & 68.33      & \underline{12.78}  & 28.96  & 94.12  \\
SIFT\cite{Lowe.2004}\tiny{IJCV'04}         & 5.41   & 19.52  & 75.01  & 2.56     & 8.48      & 72.75    & 1.24      & 4.18       & 37.58      & 8.86   & 21.55  & 87.13  \\ \midrule
RIPE \tiny{MegaDepth}        & 8.58   & 27.27  & 88.12  & \underline{3.96}     & \textbf{12.84}     & 91.07    & \underline{3.22}      & \textbf{11.78}      & 68.86       & 7.0      & 23.69  & 93.94  \\
\multicolumn{1}{c}{$\downarrow$\scriptsize{ + ACDC}}         & \gss{+1.31}   & \gss{+1.34}   & \gss{+0.34}   & \gss{+0.41}     & \rss{-0.8}      & \gss{+0.35}     & \gss{+0.14}      & \rss{-1.45}      & \gss{+1.03}     & \gss{+4.57}   & \gss{+5.37}   & \gss{+0.51}   \\
RIPE \tiny{MegaDepth+ACDC} & \underline{9.89}   & 28.61  & \underline{88.46}  & \textbf{4.37}     & 12.04     & \underline{91.42}    & \textbf{3.36}      & \underline{10.33}      & \underline{69.89}    & 11.57  & \underline{29.06}  & \underline{94.45}  \\ \bottomrule
\end{tabular}%
\caption{Evaluation RIPE on in challenging weather conditions on the Boreas dataset. The results show how RIPE can improve by incorporating data from the ACDC dataset, facilitated by our innovative training scheme. \textbf{Best} and \underline{second-best} performances are highlighted.}
\label{tab:boreas}
\end{table*}

\begin{figure*}[ht]
    \centering

    \begin{subfigure}[b]{\textwidth}
            \includegraphics[width=0.33\textwidth]{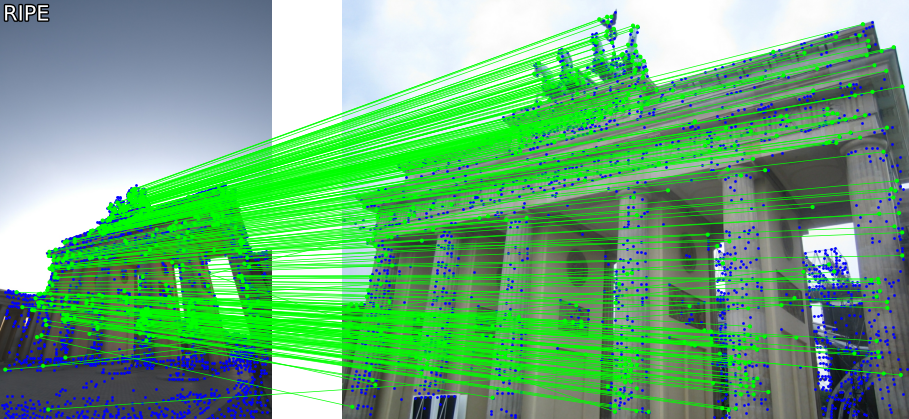}
            \includegraphics[width=0.33\textwidth]{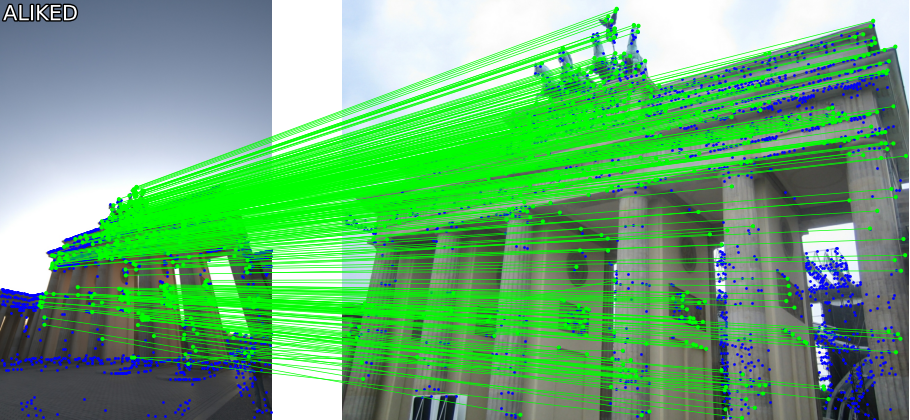}
            \includegraphics[width=0.33\textwidth]{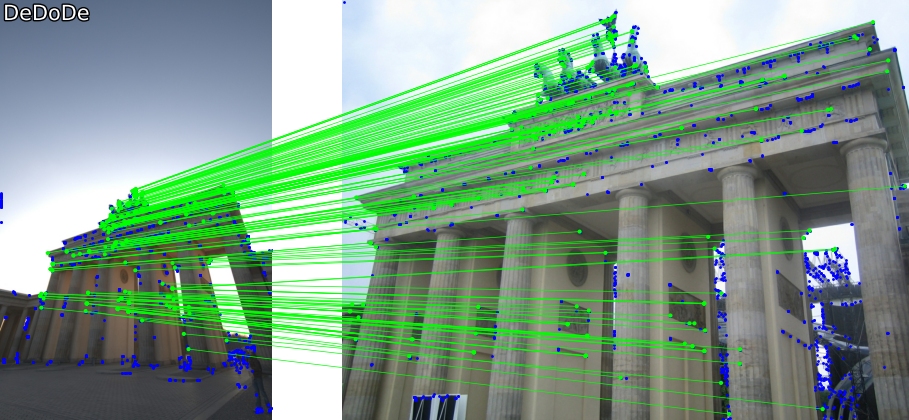}
    \end{subfigure}

    \begin{subfigure}[b]{\textwidth}
            \includegraphics[width=0.33\textwidth]{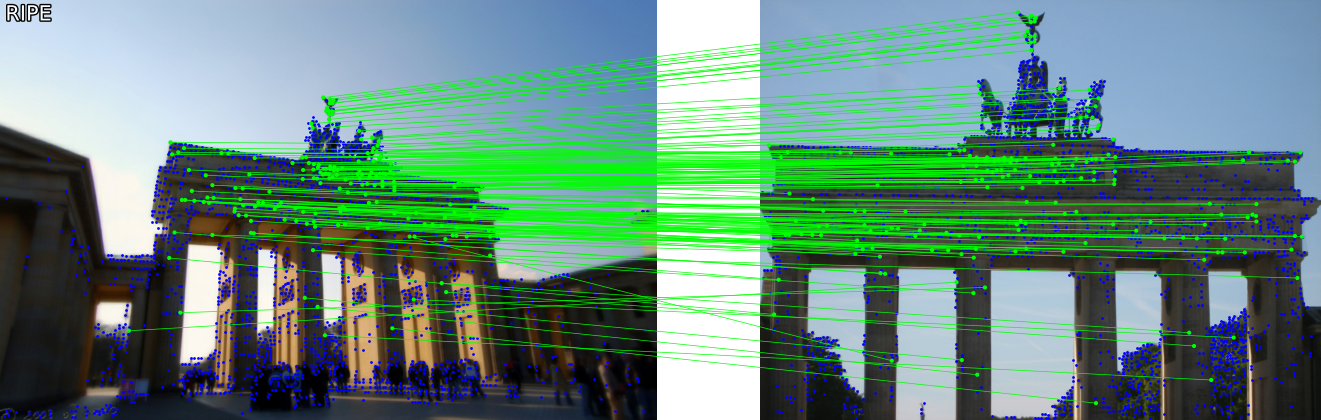}
            \includegraphics[width=0.33\textwidth]{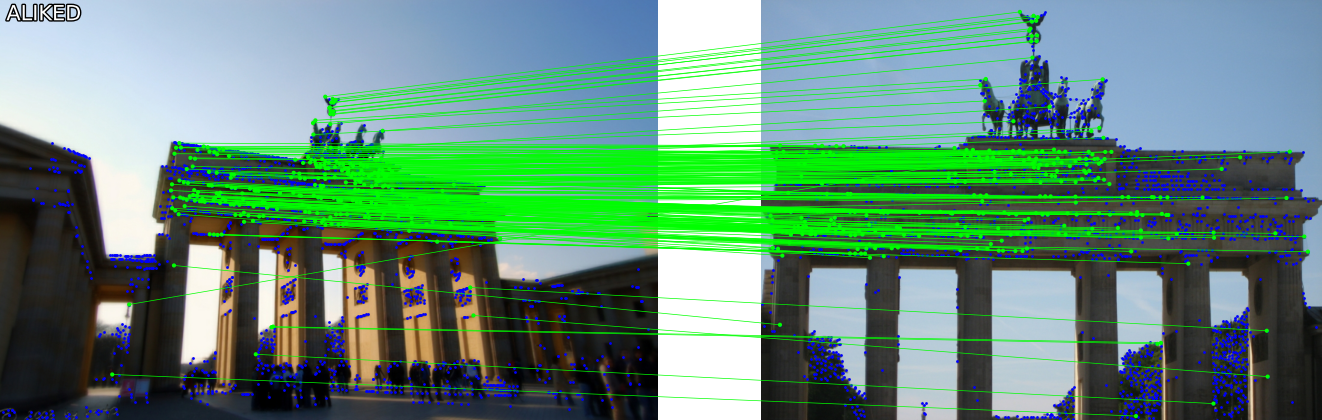}
            \includegraphics[width=0.33\textwidth]{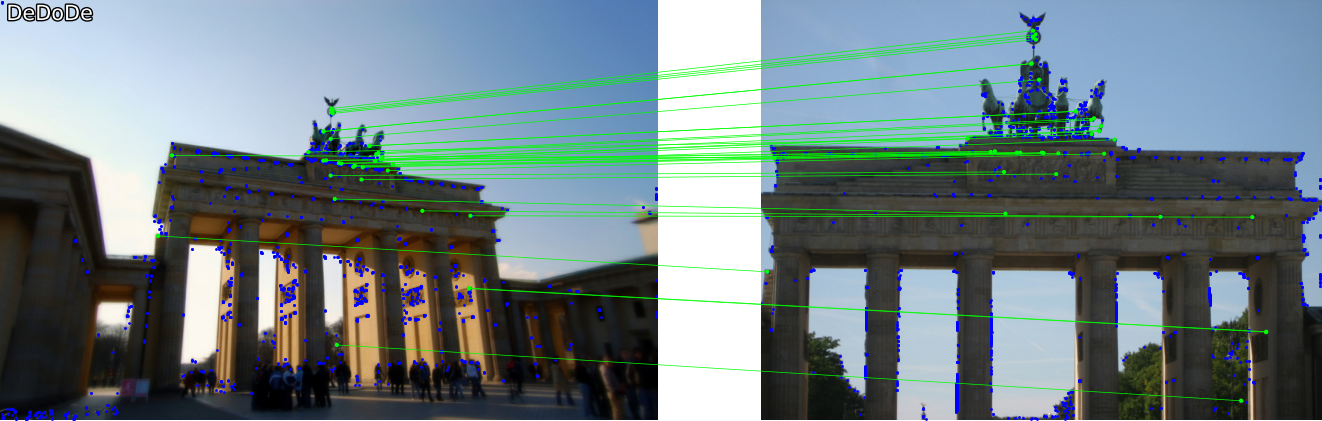}
    \end{subfigure}

    \begin{subfigure}[b]{\textwidth}
            \includegraphics[width=0.33\textwidth]{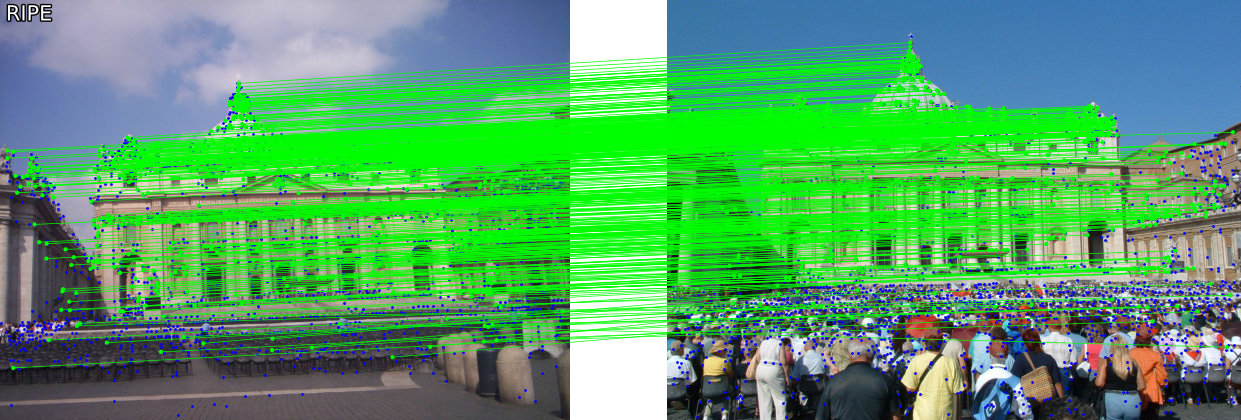}
            \includegraphics[width=0.33\textwidth]{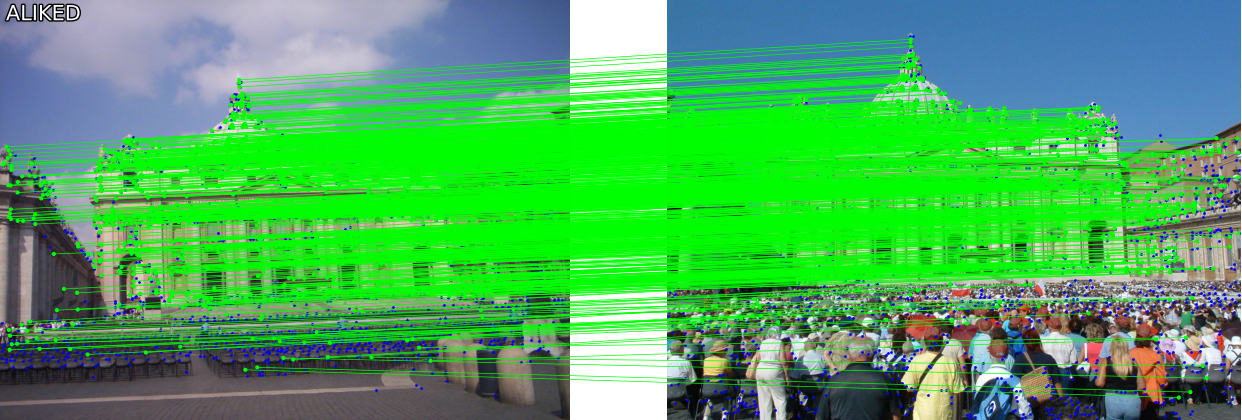}
            \includegraphics[width=0.33\textwidth]{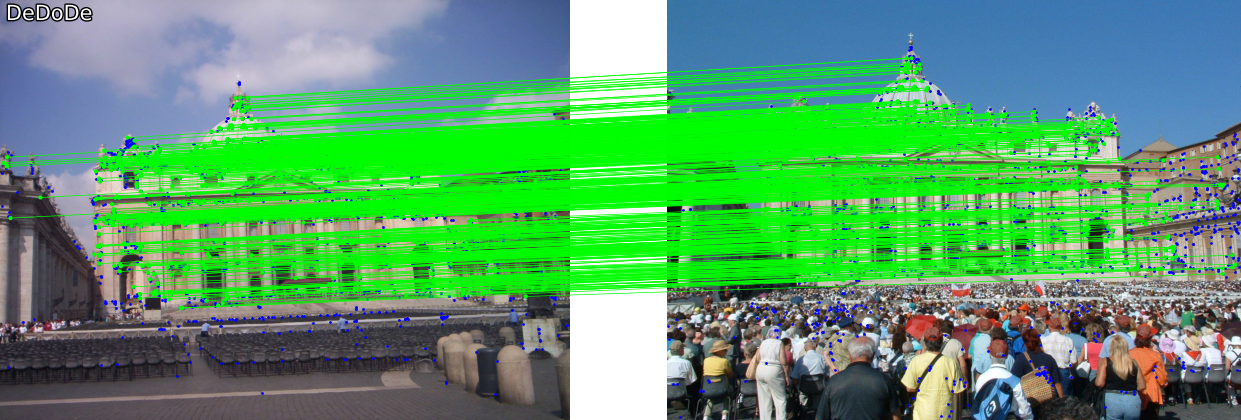}
    \end{subfigure}

    \begin{subfigure}[b]{\textwidth}
            \includegraphics[width=0.33\textwidth]{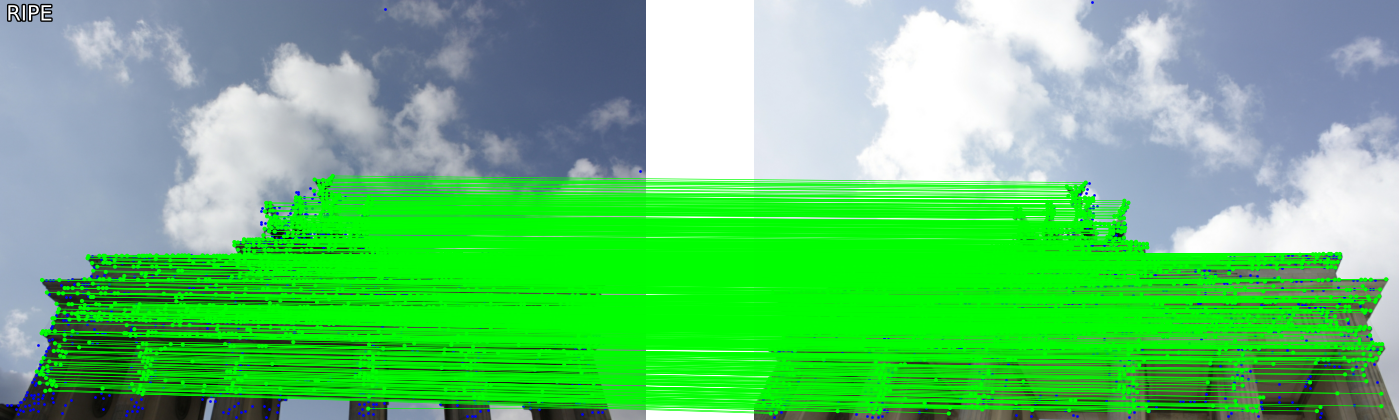}
            \includegraphics[width=0.33\textwidth]{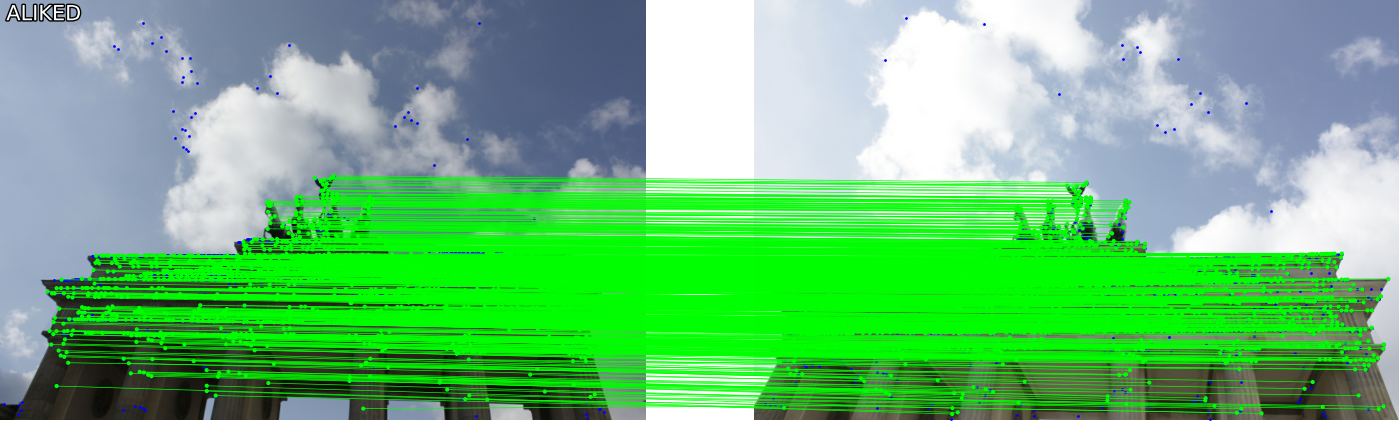}
            \includegraphics[width=0.33\textwidth]{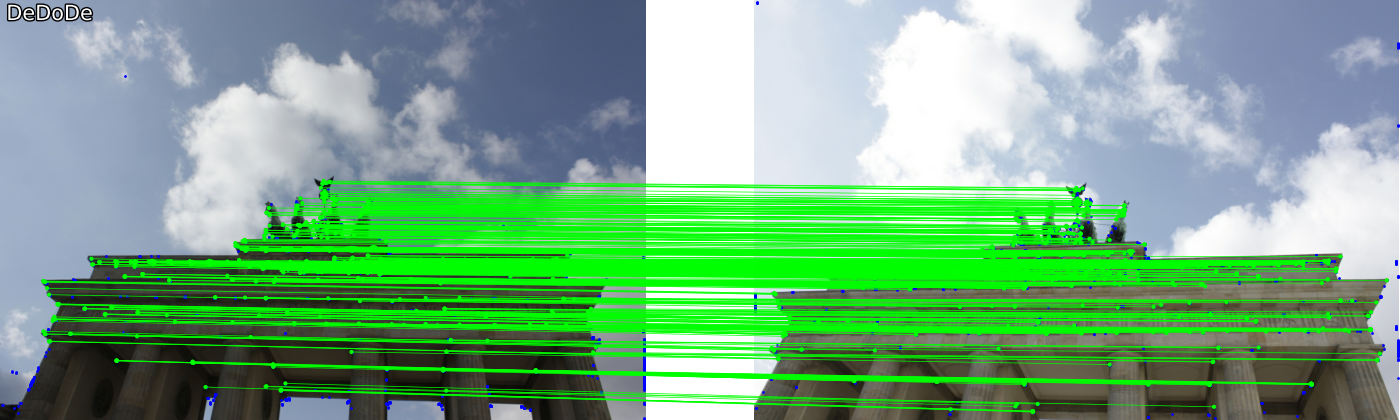}
    \end{subfigure}

    \begin{subfigure}[b]{\textwidth}
            \includegraphics[width=0.33\textwidth]{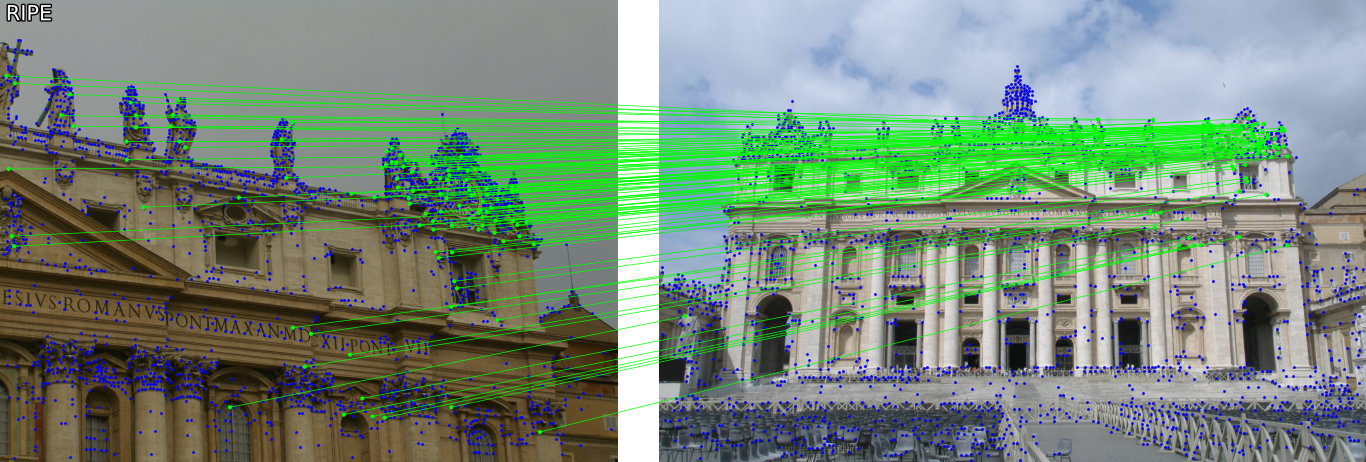}
            \includegraphics[width=0.33\textwidth]{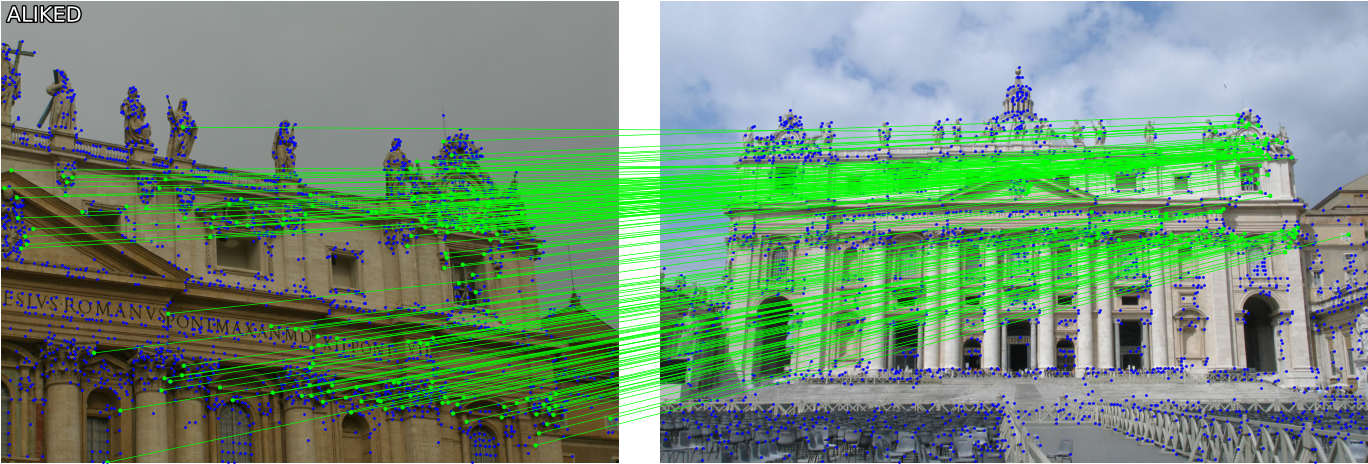}
            \includegraphics[width=0.33\textwidth]{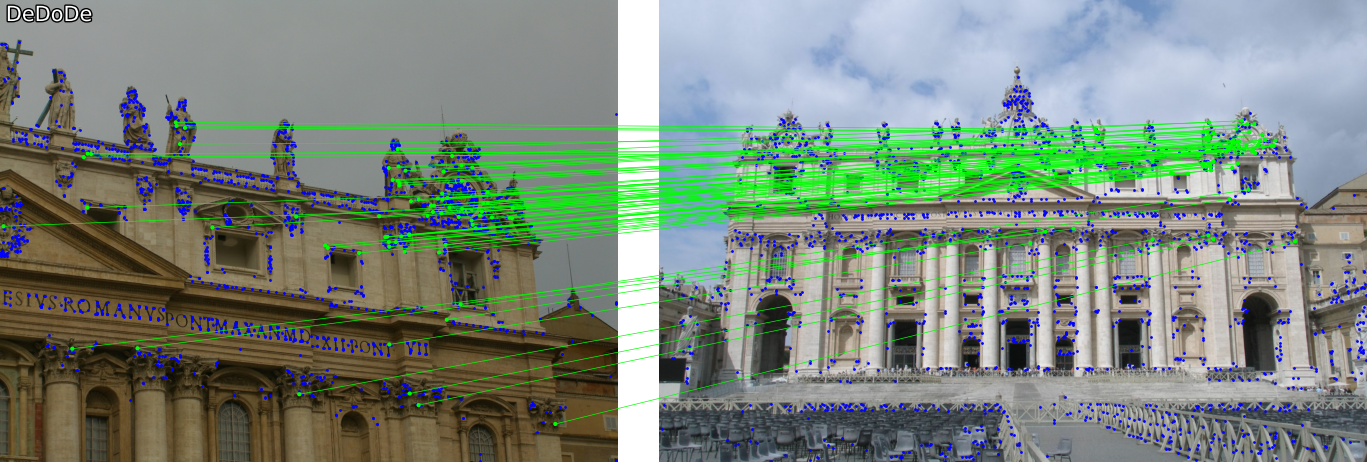}
    \end{subfigure}

    \begin{subfigure}[b]{\textwidth}
            \includegraphics[width=0.33\textwidth]{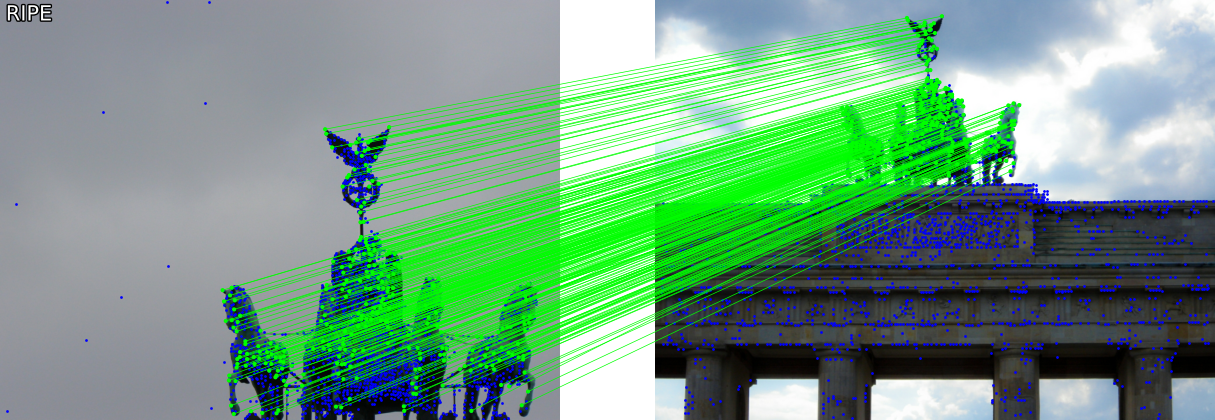}
            \includegraphics[width=0.33\textwidth]{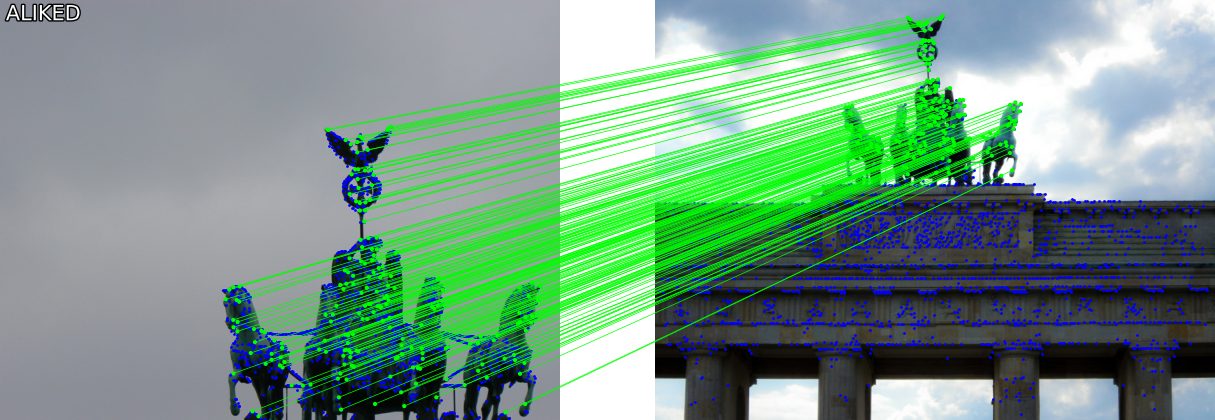}
            \includegraphics[width=0.33\textwidth]{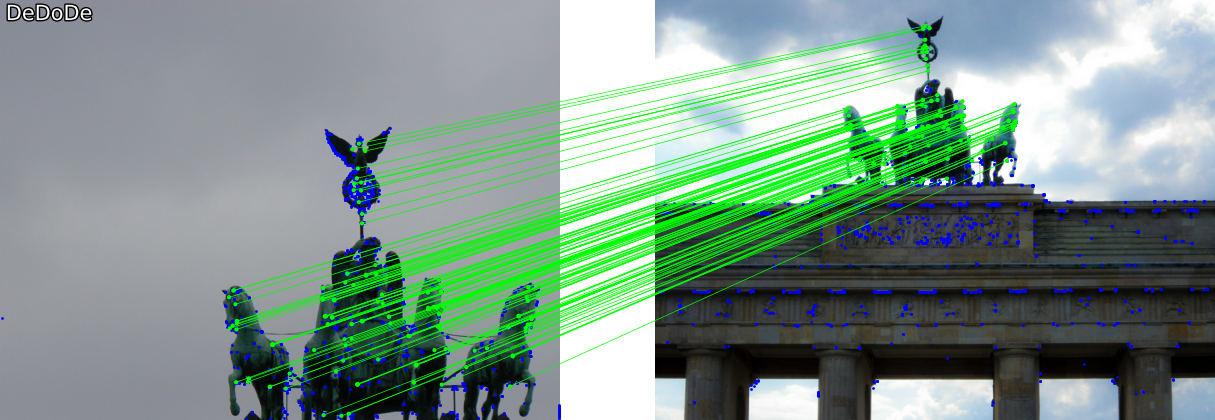}
    \end{subfigure}

%    \begin{subfigure}[b]{\textwidth}
%            \includegraphics[width=0.33\textwidth]{rsc/images/sample_3_ripe.png}
%            \includegraphics[width=0.33\textwidth]{rsc/images/sample_3_aliked.png}
%            \includegraphics[width=0.33\textwidth]{rsc/images/sample_3_dedode.png}
%    \end{subfigure}

%    \begin{subfigure}[b]{\textwidth}
%            \includegraphics[width=0.33\textwidth]{rsc/images/sample_9_ripe.png}
%            \includegraphics[width=0.33\textwidth]{rsc/images/sample_9_aliked.png}
%            \includegraphics[width=0.33\textwidth]{rsc/images/sample_9_dedode.png}
%    \end{subfigure}
    
    \caption{Example results on images from MegaDepth~1500 for RIPE (ours), ALIKED~\cite{Arandjelović.2016} and DeDoDe~\cite{Edstedt.2024}}.
    \label{fig:qualitative_result_suppl}

\end{figure*}

\subsection{Towards collapsing to the epipoles}
Our reward signal is computed based on the number of keypoints that remain after filtering for consistency with a single epipolar geometry. This raises the question of whether training could collapse by predicting keypoints only at the epipoles.

In the MegaDepth dataset, the epipoles are typically located outside the image boundaries, so this scenario does not pose a problem. In contrast, for ACDC and Tokyo~24/7, the epipoles often lie within the image area. To the best of our understanding, collapse is prevented in these cases for two reasons: first, the descriptor loss (\cref{eq:desc_loss}) promotes the learning of discriminative features; second, grid-based sampling enforces a spatially uniform distribution of keypoints during training.

In summary, collapse toward the epipoles does not occur in practice, and we never observed it in any of our experiments.

\end{document}